\documentclass[runningheads]{llncs}
\usepackage{graphicx}

\usepackage{booktabs}
\usepackage{siunitx}
\usepackage{microtype}
\usepackage{multirow}

\usepackage[utf8]{inputenc}
\usepackage{comment}
\usepackage{tablefootnote}

\usepackage{times}
\usepackage{latexsym}

\usepackage[T1]{fontenc}
\usepackage[dvipsnames]{xcolor}
\usepackage[utf8]{inputenc}
\newcommand{\squishlist}{
 \begin{list}{$\bullet$}
  { \setlength{\itemsep}{0pt}
     \setlength{\parsep}{1pt}
     \setlength{\topsep}{1pt}
     \setlength{\partopsep}{0pt}
     \setlength{\leftmargin}{1.5em}
     \setlength{\labelwidth}{1em}
     \setlength{\labelsep}{0.5em} } }
 \newcommand{\squishend}{\end{list}}

\begin{document}
\title{BigText-QA: Question Answering over a Large-Scale Hybrid Knowledge Graph\thanks{Supported by Luxembourg National Research Fund (FNR) }}
\titlerunning{BigText-QA}

\author{Jingjing Xu\inst{1}\orcidID{0000-0002-0012-3911} \and 
Maria Biryukov\inst{1}\orcidID{0000-0002-2509-5814} 
\and \\
Martin Theobald\inst{1}\orcidID{0000-0003-4067-7609} 
\and
Vinu Ellampallil Venugopal\inst{2}\orcidID{0000-0003-4429-9932}
}
\authorrunning{J. Xu et al.}

\institute{University of Luxembourg, 4365 Esch-sur-Alzette, Luxembourg \and
International Institute of Information Technology (IIIT), Bangalore, India
\email{\{jingjing.xu, maria.biryukov, martin.theobald\}@uni.lu}\\
\email{vinu.ev@iiitb.ac.in}}

\maketitle              
\begin{abstract}

 Answering complex questions over textual resources remains a challenge, particularly when dealing with nuanced relationships between multiple entities expressed within natural-language sentences. To this end, curated knowledge bases (KBs) like YAGO, DBpedia, Freebase, and Wikidata have been widely used and gained great acceptance for question-answering (QA) applications in the past decade. While these KBs offer a structured knowledge representation, they lack the contextual diversity found in natural-language sources.
To address this limitation, BigText-QA introduces an integrated QA approach, which is able to answer questions based on a more redundant form of a knowledge graph (KG) that organizes both structured and unstructured (i.e., ``hybrid'') knowledge in a unified graphical representation. Thereby, BigText-QA is able to combine the best of both worlds---a {\em canonical set of named entities}, mapped to a structured background KB (such as YAGO or Wikidata), as well as an {\em open set of textual clauses} providing highly diversified relational paraphrases with rich context information.  Our experimental results demonstrate that BigText-QA outperforms DrQA, a neural-network-based QA system, and achieves competitive results to QUEST, a graph-based unsupervised QA system.

\end{abstract}

\section{Introduction}
\label{sec:Intro}
Information extraction (IE) has made strides in extracting structured data (``facts'') from unstructured resources like text and semistructured components (e.g., tables and infoboxes)~\cite{weikum2021machine}. Established knowledge bases (KBs) such as YAGO~\cite{suchanek2007yago}, DBpedia~\cite{auer2007dbpedia}, Freebase~\cite{bollacker2008freebase} or Wikidata~\cite{vrandevcic2014wikidata} use IE techniques to store numerous facts. However, KBs are mostly limited to triple-based representations of knowledge, capturing semantic relationships between real-world objects (entities and concepts) but lacking contextual information about the facts' origins. On the other hand, information retrieval (IR) efficiently operates on large document collections using context-based statistics like term frequencies and co-occurrences~\cite{manning2009introduction}. Yet, classical IR approaches often rely on a simplified ``bag of words'' representation, overlooking the documents' internal structure.
\begin{figure}[!t]
  \centering
  \includegraphics[width=4.0in]{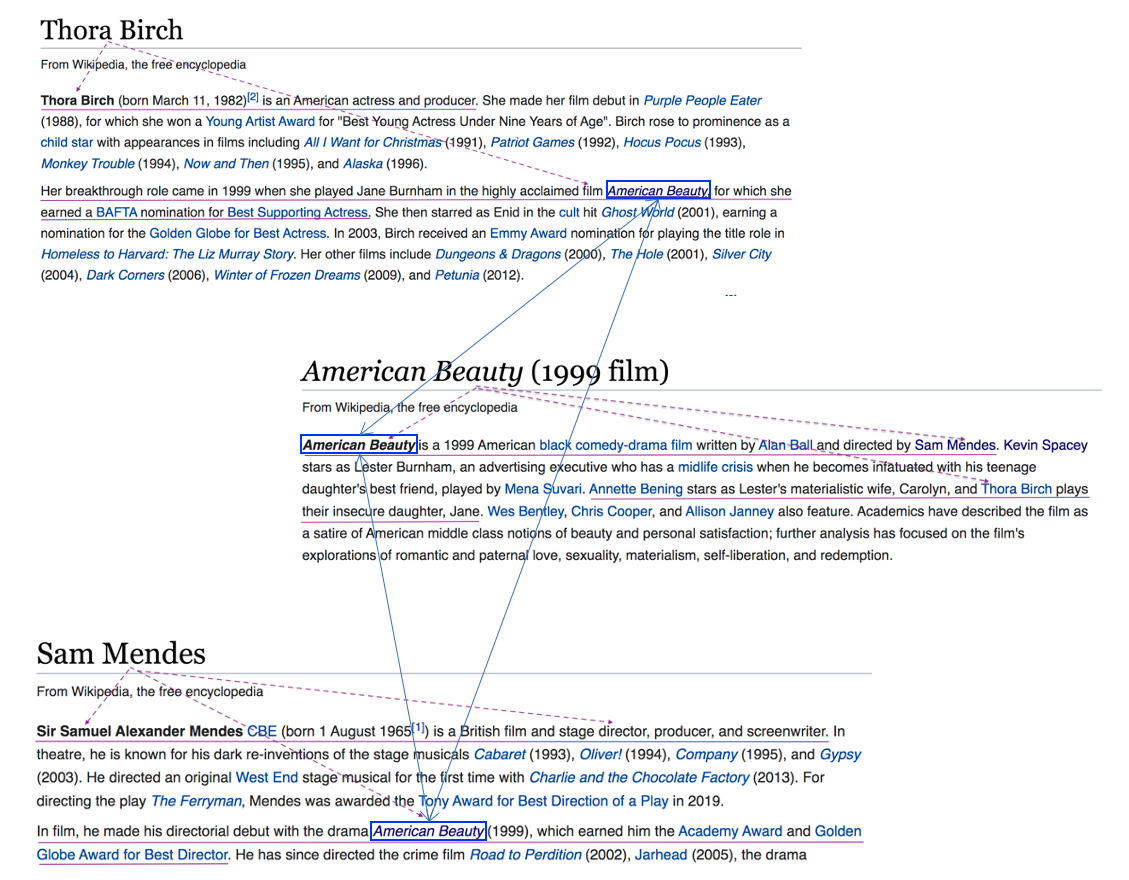}
  \caption{A snapshot of the BigText graph viewed as a corpus of interconnected documents.}
  \label{fig:BigTextDocs}
\end{figure}


Our BigText approach aims to combine the strengths of information extraction (IE) and information retrieval (IR) without simplifying their concepts. It represents a document collection as a redundant (hybrid) knowledge graph (KG), preserving the original document structure, domains, hyperlinks, and metadata. The graph includes substructures like sentences, clauses, lists, and tables, along with mentions of named entities and their syntactic and semantic dependencies. By linking mentions to a canonical set of real-world entities, additional links between entities and their contexts across document boundaries are established. Figure~\ref{fig:BigTextDocs} illustrates this approach using snapshots of Wikipedia articles about actors and movies\footnote{We use Wikipedia articles here to keep aligned with the experiments described in this paper. Note however that the choice of the document collection is not limited to any particular document type and can also combine heterogeneous natural-language resources, such as books, news, social networks, etc.}, where the articles are interconnected through jointly mentioned entities like ``American Beauty''.

From a syntactic point of view, the grammatical structures of the sentences are represented by hierarchically connected document substructures which become vertices in our BigText-KG. Specifically, {\em interlinked documents} form the basic entry points to our knowledge graph. These are decomposed into {\em sentences} which, in turn, consist of {\em clauses} (i.e., units of coherent information, each with an obligatory subject and a verbal or nominal predicate, and several optional (in)direct object(s), complement(s) and adverbials that further contextualize the two mandatory components).  
Clauses further contain {\em mentions} of {\em named entities} (NEs) within their narrow clause contexts which then finally also capture the {\em relationships} among two or more such entities.
\begin{figure}[!t]
  \centering
  \includegraphics[width=4.0in]{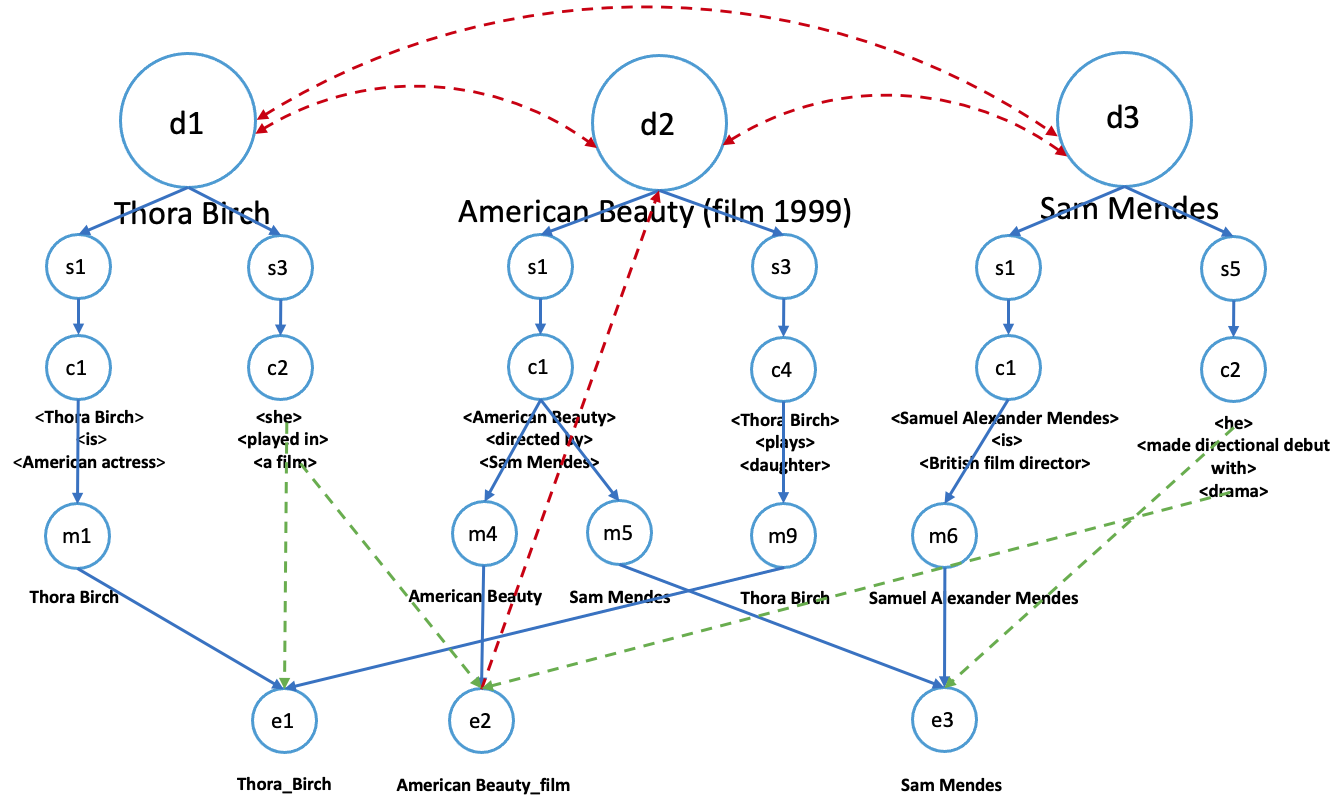} 
  \caption{Internal representation of the BigText graph of Figure~\ref{fig:BigTextDocs} as a property graph. The graph distinguishes five types of {\em vertices}: {\em documents} ($d$), {\em sentences} ($s$), {\em clauses} ($c$), {\em mentions} ($m$), and {\em entities} ($e$). The blue edges connect document's structural components. Red edges connect mentions to entities due to the named entity disambiguation and linking to real-life objects and, as a result, interlink the documents; green edges are ``implicit'' connections, linking mentions to entities due to syntactic dependencies and co-reference resolution.
  }
  \label{fig:BigTextStruct}
\end{figure}

In natural language, entities are often referred to in various ways, requiring resolution and disambiguation before mapping them to a canonical set of real-world named entities (NEs) in a background KB like YAGO or Wikidata. This process establishes reliable links within and across documents' boundaries. Syntactic dependencies are crucial for disambiguation as they expand the fixed list of vocabulary variations to dynamically occurring local contexts. For instance, consider the sentence ``\textit{she} played Jane Burnham in the highly acclaimed \textit{film American Beauty}.'' The apposition between \textit{``film''} and \textit{``American Beauty''} helps to link \textit{``she''} directly to the real-world entity \textit{``American Beauty''} instead of the generic concept \textit{``film''}. Similarly, in the sentence ``he made his directorial debut with the \textit{drama American Beauty} (1999),'' the apposition between \textit{``drama''} and \textit{``American Beauty''} enriches the entity with a semantic attribute describing its genre, which may be useful for sentiment analysis or further profiling of Sam Mendes' movie portfolio. Figure~\ref{fig:BigTextStruct} shows a small subset of explicit and implicit relations that can be established between the entities Thora Birch, Sam Mendes and American Beauty using the structural representations of the respective documents from Figure~\ref{fig:BigTextDocs}. Additionally, the analysis reveals different functions attributed to Thora Birch and Sam Mendes in American Beauty: Thora Birch \textit{plays in} ``American Beauty'' while Sam Mendes \textit{directs} ``American Beauty'' (an inverse relation obtained from a passive sentence ``American Beauty is a 1999 American black comedy-drama film written by Alan Ball and \textit{directed by} Sam Mendes.'').

By incorporating more entities, exploring the syntactic and semantic dependencies between them and connecting the mentions to their real-world concepts, BigText incrementally builds a large-scale hybrid KG of highly interlinked and semantically enriched documents. This BigText-KG is designed to serve as a generic basis for a variety of text-analytical tasks such as searching and ranking, relation extraction, and question answering.

\smallskip
In this paper, we present a case study in which BigText is employed as an underlying knowledge graph of a {\em question answering} (QA) system, BigText-QA. When evaluated on questions involving multiple entities and relations between them, BigText-QA achieves competitive results with state-of-the-art QA systems like QUEST~\cite{lu2019answering} and DrQA~\cite{chen2017reading}. The rest of the paper is organized ass follows: Section 2 provides a survey of related work; Section 3 formally presents the BigText knowledge graph; Section 4 introduces the BigText question-answering system; Section 5 presents the experimental setup; Section 6 discusses the experimental results; and Section 7 finally concludes the paper.

\section{Background \& Related Work}
In this section, we take a closer into the main QA approaches, which we broadly categorize based on their foundations: TextQA, KGQA (knowledge graph-based QA), and HybridQA (such as our BigText-QA), which seeks to integrate both textual and structured knowledge resources.

\smallskip
\noindent\textbf{TextQA.~} TextQA approaches typically retrieve answers from raw unstructured text by extracting and aggregating information from relevant documents. Early systems like START~\cite{katz2004viewing} are representative of such approaches, while more recent ones like DrQA, DocumentQA~\cite{clark2017simple}, and R3~\cite{wang2018r} employ neural-network techniques to enhance the matching capabilities. They leverage vast amounts of textual data and benefit from identifying semantic similarities but may struggle with complex queries calling for a concise structural representation. 

\smallskip
\noindent\textbf{KGQA.~} Traditionally, KG-based QA approaches transform a Natural Language (NL) input question into a logical representation by mapping NL phrases to various structured templates (e.g., in SPARQL). These templates can be executed against a query engine (e.g., by indexing RDF data)~\cite{berant2013semantic,diefenbach2018core}. Recent KGQA techniques have improved upon this approach in two main ways: (1) incorporation of information retrieval (IR)-style relaxation on the templates to allow for selection and ranking of the relevant KG subgraphs in response to the input~\cite{bordes2014question,dong2015question,hao2017end,yao2014information}; 
and (2) application of Neural Semantic Parsing (NSP) that converts the input question into a logical representation which, in turn, is translated into an actual query language understood by the KG~\cite{dong2016language,fu2020survey,lan2020query,luo2018knowledge,zhu2020knowledge}.
While KG-based QA systems are strong in handling the logical structure of the question, the KG is inherently condensed and largely oblivious to the question's context which may weaken the system's performance.

\smallskip
\noindent\textbf{HybridQA.~} 
Hybrid QA systems have been proposed for the sake of overcoming the limitations of the TextQA and KGQA paradigms while capitalizing on their respective strengths. A hybrid approach suggested in IBM seminal work~\cite{ferrucci2010building} followed by~\cite{baudivs2015modeling}, obtains candidate answers from separate structured and unstructured resources. Among the systems that adopted such an integrated approach are 
~\cite{savenkov2016knowledge}, that combines external textual data and SPARQL-based templates for answering questions; \cite{das2017question} and 
~\cite{oguz2020unik,sun2019pullnet} that employ neural networks to merge knowledge graphs (KGs) and textual resources into a common space using a universal-schema representation~\cite{riedel2013relation,verga2015multilingual}. More recent research has explored the integration of KGs into large pre-trained language models (LLMs) and their application for open-domain question answering ~\cite{ju2022grape,oguz2020unik,yasunaga2021qa,yu2021kg,zhang2023survey}. For example, UniK-QA~\cite{oguz2020unik} utilized the T5 model~\cite{raffel2020exploring}, a powerful LLM, to answer open-domain questions by leveraging heterogeneous sources.

Current state-of-the-art QA approaches achieve impressive results but also require vast amounts of data and significant time for training. Furthermore, they often need to dynamically integrate data retrieved from various external sources with KGs, adding to the systems' complexity. In contrast, BigText-QA circumvents the need for extensive training by building a unified property graph. This graph incorporates structured knowledge from disambiguated entities, while also preserving the original NL phrases that provide context and express relationships between the entities. This approach makes hybrid knowledge readily available for the QA process, thus reducing question-processing time. Furthermore, BigText-QA employs a Spark-based distributed architecture, enabling easy scalability and efficient handling of very large graphs.

 
\section{BigText Knowledge Graph}
\label{sec:BigText}

Our BigText project is driven by the strong belief that natural-language text itself is the most comprehensive knowledge base we can possibly have; it just needs to be made machine-accessible for further processing and analytics.

\smallskip
\noindent\textbf{Design \& Implementation.~} BigText aims at processing large collections, consisting of millions of text documents. We currently employ Apache Spark \cite{spark:Zaharia} and its integrated distributed graph engine, GraphX \cite{GraphX:Xin}, which allows us to model the entire collection as a unified property graph that can also be distributed across multiple compute nodes or be deployed on top of any of the common cloud architectures, if desired. 

\begin{figure}[!t]
  \centering
  \includegraphics[width=4.0in]{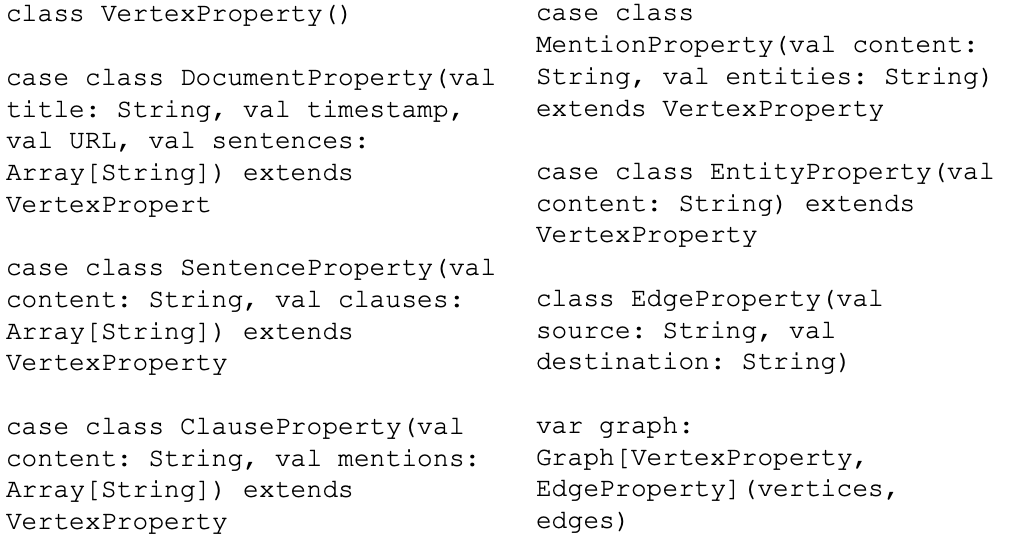}
   \caption{Case classes (in Scala) capturing the BigText-KG as a property graph in Spark's GraphX APIs.
   }
   \label{fig:VertexProp}
\end{figure}

\smallskip
\noindent\textbf{Property Graph.~} As depicted in Figure~\ref{fig:BigTextStruct}, our BigText graph distinguishes five types of {\em vertices}: {\em documents} ($d$), {\em sentences} ($s$), {\em clauses} ($c$), {\em mentions} ($m$), and {\em entities} ($e$). Spark's GraphX allows us to associate an extensible list of properties for each vertex type, such that the (ordered) vertices are able to losslessly (and partly even redundantly) capture all the extracted information from a preconfigured Natural Language Processing (NLP) pipeline together with the original text sources. Figure~\ref{fig:VertexProp} shows an internal representation of the property graph. For example, a document property stores the corresponding title and other relevant metadata, such as timestamp and source URL, while a mention vertex is augmented with morphological data, such as part-of-speech (POS) and lemma, the syntactic role within the sentence, as well as entity-type information, where applicable. Entity vertex property carries on the result of the mention disambiguation to a canonicalised entity. 

\begin{figure}[!t]
  \centering
  \includegraphics[width=4.0in]{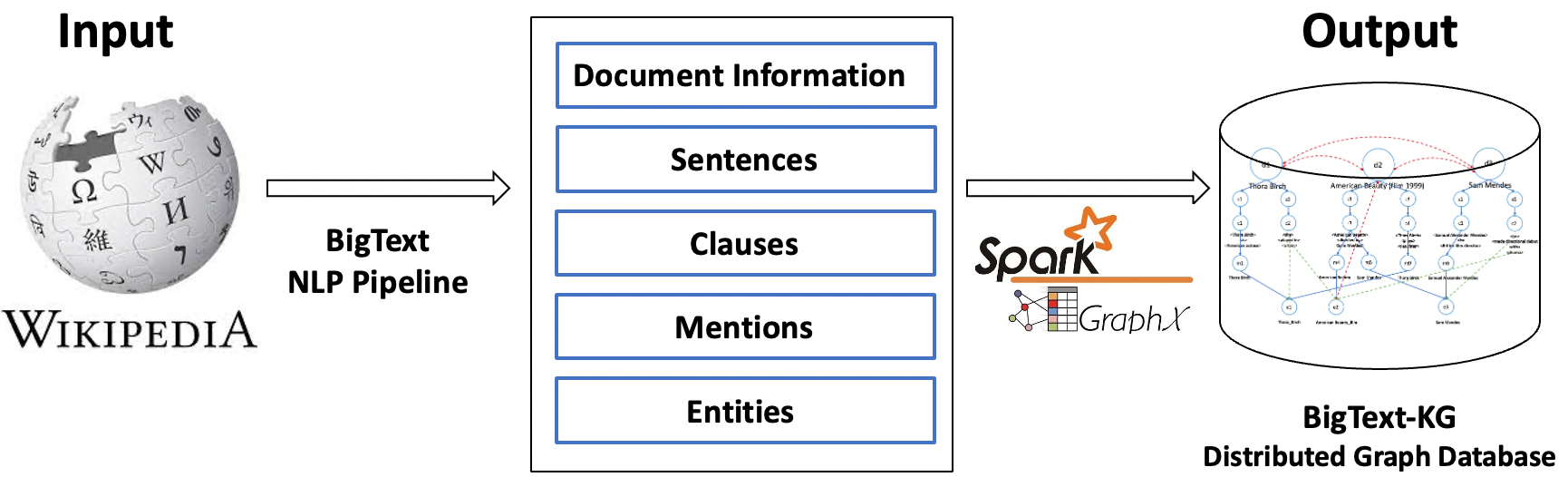}
  \caption{High-level construction of the BigText knowledge graph.}
  \label{fig:KG-pipeline}
\end{figure}

Sentences, clauses and mentions form hierarchical substructures of documents, while links among different mentions (possibly from different clauses or sentences, or even from different documents) to a same entity vertex in the background KB express additional coreferences. Recovered implicit relations resulting from appositions (e.g., ``drama'', and ``film'' with respect to ``American Beauty'') and co-reference resolution (``she'' and ``he'' with respect to Thora Birch and Sam Mendes, respectively) are shown as green thin dashed lines in Figure~\ref{fig:BigTextStruct}. Furthermore, the presence of {\em clause} vertices in combination with the disambiguated entity mentions, allows for dynamic extraction of the facts' subgraphs, containing mentions as vertices and the clauses' predicates as labeled edges.  

\smallskip
\noindent\textbf{NLP Pipeline.~} Before populating the property graph, documents in the collection are passed through a preconfigured and extensible NLP pipeline which decomposes the input into documents, sentences and clauses. Clauses are generated from sentences with an Open Information Extraction (OIE) technique. While each clause represents a semantically coherent block of entity mentions linked by a predicate, mentions first appear in their original lexical form without further linking to typed entities (such as {\tt PER}, {\tt ORG}, {\tt LOC}) or unique knowledge base identifiers, for example WikiData IDs. Therefore, our pipeline also incorporates the steps of Named Entity Recognition (NER) and Named Entity Disambiguation (NED) as they are available from recent IE tools. Since clauses may have pronouns as their subject and/or object constituents, Coreference Resolution (CR) has been added to the NLP pipeline to increase the coverage of downstream analytical tasks. For example, linking {\em she} in ``... she played Jane Burnham in the highly acclaimed film American Beauty'' to {\em Thora Birch} establishes a connection between the two real-world entities,  ``American Beauty'' and ``Thora Birch'', which can then further be explored. 
\begin{table}[htb]
    \centering
    \scriptsize
    \caption{Annotators and background KBs used in the BigText NLP pipeline.}
    \label{tab:annotators}
    \begin{tabular}{c|c}
    \toprule
    \textbf{Annotation type} & {\textbf{Tools}}  \\
    \midrule
    HTML parser & \textbf{Jsoup}\tablefootnote{\scriptsize\url{https://jsoup.org/}} \\
     \midrule
    Tokenization  & Spacy  \\
     \midrule
    OpenIE & \textbf{ClausIE}~\cite{CIE},  OpenIE5\tablefootnote{\scriptsize\url{https://github.com/dair-iitd/OpenIE-standalone}}, OpenIE6~\cite{OIE6} \\
     \midrule
    NER & \textbf{StanfordNLP}, Flair~\cite{Flair}\\
     \midrule
    NED & \textbf{AIDA-Light}~\cite{nguyen2014}, REL~\cite{REL}, ELQ~\cite{ELQ} \\
        \midrule
    CR & SpanBERT:2018~\cite{LeeSB}, SpanBERT:2020~\cite{JoshiSB}\\
     \midrule
     Background KB & \textbf{YAGO}, WikiData\tablefootnote{\scriptsize\url{https://www.wikidata.org/}}\\
    \bottomrule
    \end{tabular}
\end{table}

Projects which involve the stage of text (pre-)processing typically apply either an entire end-to-end suite of annotation tools, such as 
NLTK~\cite{bird2009natural}, 
StanfordNLP~\cite{manning2014stanford}, 
SpaCy\footnote{\scriptsize\url{https://spacy.io/}}, or a specific component from it (which can also be substituted with a stand-alone or equivalent tool). Conversely, our text annotation pipeline does not limit the choice of annotators. We intend to use state-of-the-art target-specific components to minimise the risk of error propagation. This strategy allows us to adapt the selection of tools to the type of documents being processed (e.g., long documents corresponding to full-text Wikipedia articles versus short ones, such as Wikipedia articles' abstracts or news). Our implementation also allows for integrating outputs provided by different tools with the same annotation goal. In that way, the pipeline can be configured with further rules that prioritize either precision or recall (e.g., by considering either the intersection or the union of annotations). Table \ref{tab:annotators} depicts the annotation tools that have been integrated into the BigText NLP pipeline so far. Figure~\ref{fig:KG-pipeline} depicts the entire construction process of the BigText knowledge graph (BigText-KG).

\begin{table}[htb]
  \centering
  \caption{BigText-KG statistics (in millions) for Wikipedia.}
  \label{tab:graph_stats}
  \begin{tabular}{ccccc}
  \toprule
  Documents& Sentences& Clauses& Mentions & Entities \\
  \midrule
  5.3 & 97 &190 &283 &2 \\
  \bottomrule
  \end{tabular}
\end{table}

\smallskip
\noindent\textbf{Applications.~} In the following part of this paper, we focus on {\em question answering} (QA) as our main target application which relies on the BigText-KG as its underlying knowledge graph. We use full-text articles of an entire Wikipedia dump from 2019\footnote{\scriptsize\url{https://dumps.wikimedia.org/enwiki/latest/}}. Statistics are shown in Table \ref{tab:graph_stats}. Tools that have been used to process the version discussed here and used for the experiments are shown in Table \ref{tab:annotators} in bold font. 

\section{BigText Question Answering}
\label{sec:bigTextQA}
The design of BigText-QA is based on QUEST, a graph-based question-answering system that specifically targets complex questions with multiple entities and relations. QUEST constructs a so-called {\em quasi-graph} by ``googling'' for relevant documents in response to an NL input question and by applying a proximity-based decomposition of sentences into {\tt <sub>}, {\tt <pred>}, {\tt <obj>} triplets (SPO). Similarly to our BigText-KG, leaf nodes of the quasi-graph are {\em mentions} (vertex labels in BigText), {\em relations} (edge properties in BigText) and {\em type} nodes (vertex properties in BigText). In QUEST, the latter are the result of a semantic expansion of mentions via the application of Hearst patterns~\cite{hearst1992automatic} and/or lookups in an explicit mention-entity dictionary. In BigText, both the structural decomposition of the documents and their annotations have been provided by the preconfigured NLP pipeline. 

Since our instance of the BigText-KG is built using Wikipedia as the text resource, our system consequently retrieves relevant Wikipedia documents by using Lucene as underlying search engine\footnote{\scriptsize\url{https://lucene.apache.org/core/}}. The top-10 of the retrieved documents then serve as pivots for the respective subgraph that is selected from the entire BigText-KG upon each incoming NL question. In summary, we translate such a BigText subgraph to a structure equivalent to QUEST's quasi-graph (depicted in Figure \ref{fig:quasi-graph}) as follows:

\smallskip
\noindent\textbf{Vertex translation.~}  Mention (``m'') and entity (``e'') vertices are directly translated from the BigText subgraph to the QUEST quasi-graph. These can be the subject and/or object of a clause. Type vertices (``t'') are added based on the syntactic and semantic properties of the vertices, augmented with the application of Hearst patterns, following the QUEST approach.
Predicate vertices (``p'') are created out of the verbal component of a clause, while synonymous relation nodes are added using word/phrase embeddings (Section \ref{sec:QA_pipeline}).  

\smallskip
\noindent\textbf{Edge translation.~} Edges between predicates, mentions and disambiguated entities are directly translated from the BigText subgraph into the QUEST quasi-graph. Similarly to QUEST, we additionally introduce \emph{type} and \emph{alignment} edges from the respective vertex and edge properties in BigText. Type edges connect mention nodes with type nodes. For example, an organge edge between the mention node ``Thora Birch'' and type node ``American actress'' on the Figure \ref{fig:quasi-graph} is one such edge. It expresses the relation of type \emph{$NP_1$ is $NP_2$} captured by Hearst pattern. Alignment edges connect potentially synonymous mention nodes resolved to the same entity (thick blue edges between, for example, $m_5$ ``Sam Mendes'' and $m_6$ ``Samuel Alexander Mendes'', connected via $e_3$ to the entity ``Sam\_Mendes''), and potentially synonymous relation nodes such as ``made directional debut'' and ``directed by''(dashed blues edges) in the same figure.

\subsection{Question-Answering Pipeline}
\label{sec:QA_pipeline}
In more detail, our QA pipeline processes an incoming NL question (or clue) as follows.
\squishlist
\item [(1)] The NL input question serves as a keyword query to Lucene which then retrieves the top-10 most relevant documents from the Wikipedia corpus.
These documents are used as entry points to the BigText-KG to select a relevant subgraph that captures the questions' context; this subgraph includes all the documents' hierarchical substructures plus their links to the background KB. 
\item [(2)] A syntactic parser (similar to QUEST) is applied to the question in order to identify its subject, predicate and object, which we refer to as the \textit{question terms}. 
\item [(3)] The subgraph is translated into QUEST's quasi-graph (see below for details).
\item [(4)] Vertices in the quasi-graph, which have high similarity with the question terms, become terminals (so-called ``cornerstones'' in QUEST).
For example, ``Thora Birch'', ``played in'' and ``plays'' are examples of cornerstones, corresponding to question terms ``Thora Birch'' and ``starred'' (orange nodes in Figure \ref{fig:quasi-graph}), respectively.
\item [(5)] Together with the quasi-graph and its weighted edges (see below), cornerstones constitute the input to QUEST's Group Steiner Tree (GST) algorithm~\cite{GST} which is used to compute an answer set. A final ranking among the matching vertices in the answer set provides the ranked answers to the input question.
\squishend

\begin{figure}[!t]
  \centering
  \includegraphics[width=4.0in]{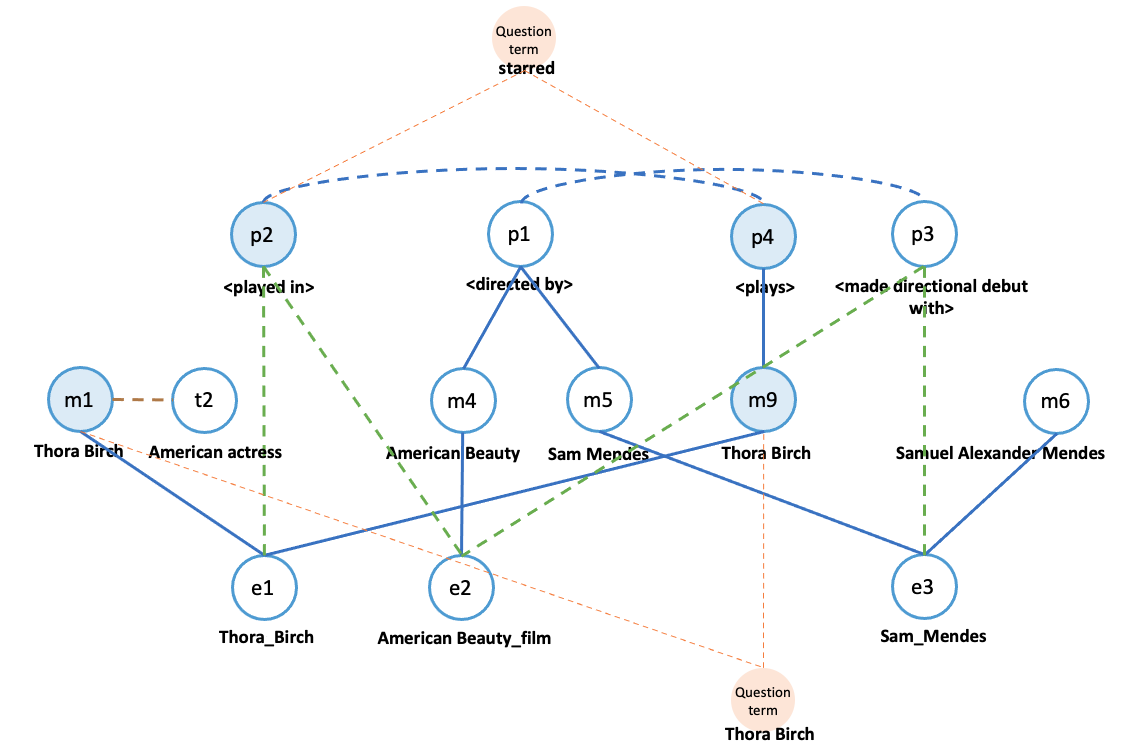}
  \caption{QUEST quasi-graph for the question \emph{``Which British stage director is best known for his feature-film directing debut, which starred Kevin Spacey, Annette Bening, and Thora Birch?''}. It results from the translation of the BigText (sub)graph in Figure \ref{fig:BigTextStruct}. }
  \label{fig:quasi-graph}
\end{figure}

\subsection{Weighting Schemes}
\label{sub:weighting}
Before we can proceed with the application of the GST algorithm and its answer-set calculation, both vertices and edges in QUEST's quasi-graph have to be assigned with weights, which (in our case) are derived from the relevant BigText subgraph.
\squishlist
\item\textbf{Vertex weights} are defined by the similarity between the question terms and the vertices in the quasi-graph.
Specifically, we adopt the two weighting schemes suggested in~\cite{lu2019answering} and discuss their application.
\item\textbf{Edge weights} are calculated depending on the edge type. The weight of an edge between a mention and a predicate vertex is the inverse of the distance between the two vertices in the BigText subgraph. Formally, this is defined as the number of words between a mention (i.e., subject and object) and a predicate of a clause vertex in the BigText subgraph. If two vertices are directly connected via multiple edges, the highest such weight is selected.
Weights of alignment edges are calculated based on the semantic similarity between the vertices they connect (see Subsection~\ref{sub:similarities}).
\squishend

\subsection{Similarities \& Thresholds}
\label{sub:similarities}
Question terms are compared to the vertices in the quasi-graph according to their syntactic type as follows.

\smallskip
\noindent\textbf{Jaccard Similarity.~} The AIDA dictionary~\cite{hoffart2011robust} is a dictionary composed of a large number of entity-mention pairs in YAGO, where each mention is associated with the set of entities it may refer to; entities are represented by their unique identifiers. The ``subject'' and ``object'' question terms (see Section \ref{sec:QA_pipeline}) are compared to the mention nodes using Jaccard similarity. Similarly, Jaccard similarity is between the entity sets associated with a mention vertex and the question term extracted from the input question. For mentions and question terms that could not be found in the AIDA dictionary, the Jaccard similarity is computed based on the plain string similarity between the two. In either case, the maximal value between a mention/entity and a question term selected as the vertex weight in the quasi-graph.

\smallskip
\noindent\textbf{Cosine Similarity.~} Question terms identified as ``predicates'' are compared to the {\em predicate nodes} of the quasi-graph in a pair-wise manner using the Cosine similarity between their corresponding word embeddings\footnote{We use the default word2vec model~\cite{mikolov2013distributed} trained on Google news.}. For each predicate node in the quasi-graph, the maximal cosine similarity value of all pair-wise comparisons is selected as its weight.

\smallskip
\noindent\textbf{Similarities for Alignment Edges.~} Once the node weights of the quasi-graph are computed, we can decide on the insertion of additional {\em alignment edges} into QUEST's quasi graph to further support the GST algorithm. An alignment edge is inserted if the similarity between two candidate vertices of the same type exceeds a pre-defined threshold. The similarity value then becomes the weight of the corresponding edge. Similarities between two mention vertices are again computed using Jaccard similarity (otherwise it is set to $1$ if the two mentions are linked to the same entity), while the ones between predicate and type vertices are calculated using Cosine similarity.

\smallskip
\noindent After the quasi-graph has been constructed, it is its largest connected component, together with the cornerstones, which is used as an input to the GST-algorithm. 

\smallskip
\noindent\textbf{Thresholds.~} 
All thresholds (calculated either by Jaccard or Cosine similarities) are set to 0.25 except the ones for the \emph{predicate alignment edges} where we experiment with a range of values: 0.25, 0.375, 0.5, 0.6, 0.75 (see Table \ref{Acc-Ans-111}). We remark that our threshold policy is different from the one applied by QUEST, where all thresholds are the same and set to 0.5.

\smallskip
After the quasi-graph construction, we proceed with the answer-set computation, ranking and filtering. These steps are performed exactly as in the QUEST framework.

\section{Experiments}
\label{sec:set}
All experiments reported in this paper are conducted on a single large Intel Xeon Platinum server with 2.4 GHz, 192 virtual cores and 1.2 TB of RAM, holding the entire BigText-KG in main memory. All translation steps are performed using PySpark 2.4.1 \cite{zaharia2016apache} for transforming Spark's GraphX RDDs into the relevant BigText subgraph via parallel processing. The BigText subgraph is translated into QUEST's quasi-graph by a second Python library, NetworkX 2.8 \cite{hagberg2008exploring}.

We use two benchmark datasets for the evaluation:
CQ-W~\cite{abujabal2017automated} and TriviaQA~\cite{joshi2017triviaqa}. Regarding CQ-W, we remove questions whose answers are not present in the Wikipedia-based BigText subgraph, which is the case for about 25\% of the questions\footnote{This decision was motivated by the fact that, for those questions, none of the top-10 documents returned by Lucene actually contained the answer.}. The remaining 75\% of the questions are used for the comparative evaluation. As for TriviaQA, we randomly select 79 questions from the development set ({\tt wikipedia-dev.json}). CQ-W is a curated dataset of question-and-answer pairs, which consists of 150 complex questions from Wiki-Answers~\cite{fader2013paraphrase}. TriviaQA is a large-scale dataset made of complex and compositional questions and corresponding gold answers.

To run the experiments, we feed the top-10 documents selected from Wikipedia by Lucene both to the original QUEST engine and to BigText-QA in order to ensure a fair comparison. 
We also quote the results achieved by DrQA on the CQ-W dataset as a further baseline. As opposed to QUEST and BigText-QA, DrQA is a neural-network-based QA system, and thus represents another class of QA systems. DrQA is trained on the SQuAD~\cite{rajpurkar2016squad} question-and-answer set which is also based on a subset of Wikipedia articles. 
\section{Result \& Discussion}
\label{sec:res}
\begin{table}
  \scriptsize
  \centering
  \caption{Comparison between BigText-QA, QUEST and DrQA on the CQ-W and TriviaQA datasets.}
  {
  \begin{tabular}{ccccccrr}
  \toprule
  \textbf{Dataset}        & \textbf{System}             & \textbf{Cosine}    &\textbf{\#Vertices} & \textbf{\#Edges ($10^5$)}   & \textbf{MRR}   & \textbf{P@1}  & \textbf{Hit@5}    \\
  \midrule
                          & \multirow{5}{*}{BigText-QA}    & 0.250                    & 1,276           & 7.234           & 0.387          & 0.324          & 0.441          \\
                          &          & 0.375                    & 1,276           & 7.234           & 0.387          & 0.324          & 0.441                \\
                          &                             & \textbf{0.500}     & \textbf{1,268}        & {6.727}      & \textbf{0.398} & \textbf{0.342} & \textbf{0.423}  \\
                          &                             & \textbf{0.600}     & \textbf{579}          & {0.510}      & \textbf{0.264} & \textbf{0.198} & \textbf{0.297}   \\
\multirow{3}{*}{CQ-W}     &                             & 0.750                    & 210             & 0.030          & 0.140          & 0.081          & 0.189  \\
                  \cmidrule(r){2-8}
                          &\multirow{3}{*}{QUEST}       & 0.500                    & 2,385           & 13.580          & 0.464          & 0.423          & 0.495                 \\
                          &                             & \textbf{0.600}     & \textbf{1,267}        & {0.609}        & {0.329}        & {0.279}        & {0.369}       \\
                          &                             & \textbf{0.750}     & \textbf{642}          & {0.032}        & {0.181}        & {0.099}        & {0.279}    \\
                  \cmidrule(r){2-8}
                          & DrQA                        & -                  & -               & -                   & {0.120} & {0.171} & {0.315}   \\
                  \midrule
                          & \multirow{4}{*}{BigText-QA}    & 0.375              & 840             & 2.073               & 0.412          & 0.342          & 0.494                 \\
                          &                             & \textbf{0.500}     & \textbf{838}    & {1.968}      & \textbf{0.412}     & \textbf{0.342}    & \textbf{0.468}   \\
\multirow{2}{*}{Trivia-QA}   &                             & \textbf{0.600}     & \textbf{365}    & {0.163}      & \textbf{0.258}     & \textbf{0.190}    & \textbf{0.316}     \\
                          &                             & 0.750              & 121             & 0.007               & 0.130          & 0.063          & 0.190                   \\
                  \cmidrule(r){2-8}
                          & \multirow{3}{*}{QUEST}      & 0.500              & 1,710           & 4.025               & 0.425          & 0.380          & 0.468               \\
                          &                             & \textbf{0.600}     & \textbf{968}    & {0.241}             & {0.285}        & {0.215}        & {0.329}     \\
                          &                             & \textbf{0.750}     & \textbf{490}    & {0.025}             & {0.198}        & {0.139}        & {0.241}    \\
  \bottomrule
  \end{tabular}}
  \label{Acc-Ans-111}
\end{table}
The main scoring metric for evaluating the QA systems is Mean Reciprocal Rank (MRR), calculated as $MRR = {\frac{1}{Q}} {\sum_{i=1}^{Q} \frac{1}{rank_{i}}}\label{MRR}$, where Q is the number of questions and $rank_i$ is the rank of the first correct answer for the $i$-th question. Other important metrics include Precision@1 (P@1) and Hit@5. P@1 represents the precision of the top-ranked document retrieved, while Hit@5 is 1 if one of the top 5 results includes the correct answer. These metrics offer valuable insights into the system's performance. A higher score indicates better performance for the QA system. The results are shown in Table \ref{Acc-Ans-111} and Table \ref{Result-Metric-Types}. In these tables, \emph{Cosine} refers to the edge threshold which is used to select \emph{predicate alignment edges}; \emph{\#vertices} and \emph{\#edges} refer to the largest connected component of the quasi-graph which is used as input to the GST-algorithm. \emph{MRR}, \emph{P@1}, \emph{Hit@5} (as well as the number of nodes and edges in the respective quasi-graphs) are averaged across all questions. 

Table \ref{Acc-Ans-111} shows that BigText-QA achieves very competitive results, usually generating more compact yet denser graphs compared to QUEST. This compactness positively impacts the results, as it involves fewer answer candidates. QUEST slightly outperforms BigText-QA when its quasi-graph has nearly twice as many nodes. However, in cases where both systems generate quasi-graphs of comparable order (BigText-QA with $\mathit{Cosine} = 0.5$, QUEST with $\mathit{Cosine} = 0.6$, and BigText-QA with $\mathit{Cosine} = 0.6$, QUEST with $\mathit{Cosine} = 0.75$, shown in bold font in Table \ref{Acc-Ans-111}), BigText-QA outperforms QUEST on both question sets, CQ-W, and TriviaQA.

Table \ref{Acc-Ans-111} indicates two ways of achieving close results using the GST algorithm: either by having sufficient, even ``poorly'' connected vertices (QUEST), or by having fewer but better connected vertices (BigText-QA). Different quasi-graph configurations in both systems arise from distinct underlying NLP pre-processing of input documents, particularly in the decomposition of sentences into clauses. BigText-QA yields fewer but more accurate vertices, facilitating the generation of dense graphs even with the increasing threshold values for edge insertion, which positively affects the performance.

\begin{table}
  \scriptsize
  \centering
  \caption{Comparison of QUEST and BigText-QA over different categories of questions in CQ-W. }
  {
  \begin{tabular}{ccccccrr}
  \toprule
  \textbf{Type}     & \textbf{System}  & \textbf{Cosine}  &\textbf{\#Vertices} &\textbf{\#Edges($10^5$)} & \textbf{MRR}     & \textbf{P@1}   & \textbf{Hit@5}   \\
\midrule
                      &\multirow{3}{*}{BigText-QA}     & \textbf{0.500}   & \textbf{1,375}  &{9.451}           & \textbf{0.388}   & \textbf{0.333} & \textbf{0.407} \\
                      &         & \textbf{0.600}   & \textbf{654}    & {0.667}          & \textbf{0.273}   & \textbf{0.185} & \textbf{0.333} \\
{People}              &                             & 0.750            & 250             & 0.035            & 0.173            & 0.111          & 0.204          \\ 
                 \cmidrule(r){2-8} 
                      & \multirow{3}{*}{QUEST}      & 0.500            & 2,600           & 19.396           & 0.448            & 0.389          & 0.500          \\
                      &                             & \textbf{0.600}   & \textbf{1,384}  & {0.793}          & {0.304}          & {0.259}        & {0.333} \\
                      &                             & \textbf{0.750}   & \textbf{741}    & {0.037}          & {0.159}          & {0.074}        & {0.259} \\ 
\midrule
                      & \multirow{3}{*}{BigText-QA}    & \textbf{0.500}   & \textbf{1,428}  & {4.843}          & {0.353}          & {0.333}        & {0.333} \\
                      &          & \textbf{0.600}   & \textbf{657}    & 0.507            & 0.242            & 0.200          & 0.233          \\
{Movie}               &                             & 0.750            & 185             & 0.029            & 0.074            & 0.033          & 0.100          \\ 
                \cmidrule(r){2-8}  
                      & \multirow{3}{*}{QUEST}      & 0.500            & 2,636           & 9.279           & 0.504            & 0.500           & 0.500          \\
                      &                             & \textbf{0.600}   & \textbf{1,381}  & 0.496           & \textbf{0.441}   & \textbf{0.433}  & \textbf{0.433} \\
                      &                             & \textbf{0.750}   & \textbf{589}    & 0.027           & \textbf{0.094}   & \textbf{0.067}  & \textbf{0.067} \\ 
\midrule
                      &\multirow{3}{*}{BigText-QA}     & \textbf{0.500}   & \textbf{756}    & 2.059           & \textbf{0.580}   & \textbf{0.444} & \textbf{0.722} \\
                      &          & {0.600}          & {254}           & 0.107           & {0.293}          & {0.167}        & {0.389}    \\
{Place}               &                             & 0.750            & 129             & 0.006           & 0.108            & 0.056          & 0.222          \\ 
                \cmidrule(r){2-8} 
                      & \multirow{3}{*}{QUEST}      & 0.500            & 1,393           & 3.126           & 0.498            & 0.444          & 0.556          \\
                      &                             & \textbf{0.600}   & \textbf{729}    & {0.168}         & {0.315}          & {0.167}        & {0.500} \\
                      &                             & 0.750            & 414             & 0.013           & 0.359            & 0.222          & 0.500          \\ 
  \bottomrule
  \end{tabular}}
  \label{Result-Metric-Types}
\end{table}

Note, that the effect of changing the edge threshold below 0.5 is negligible (first two rows in the Table \ref{Acc-Ans-111}). However, raising it from 0.5 to 0.6 significantly decreases the number of edges. This may be due to the word2vec model itself, as a low similarity threshold results in numerous ``weak'' alignment edges. A change from 0.5 to 0.6 is typically where the model becomes more discriminative, leading to a smaller number of synonymous edges. This trend continues with a further increase in the threshold to 0.75.

When compared to DrQA, both QUEST and BigText-QA outperform it, as they are designed to handle complex questions and incorporate evidence from multiple documents. In contrast, DrQA embodies an IR-based approach to QA, expecting the answer to a question to be narrowed down to a specific text span within a single document that closely matches the question.


To gain a more detailed understanding of the performance of BigText-QA versus QUEST, we divided the CQ-W set into six question categories: \textit{People}, \textit{Movie}, \textit{Place}, \textit{Others}, \textit{Language}, \textit{Music}. However, the last three categories turned out too small to be representative (containing only 4, 2 and 3 questions, respectively). We therefore focus on the first three categories (\textit{People}, \textit{Movie}, \textit{Place}) in Table \ref{Result-Metric-Types}. Here again we highlight in boldface the lines which show the results obtained by both the systems on the quasi-graphs of comparable order. Consistent with the overall results, BigText-QA and QUEST demonstrate similar performance patterns in \textit{People}- and \textit{Place}-related questions as we discussed previously. However, BigText-QA does not compare favorably in \textit{Movie}-related questions. We leave an in-depth investigation of this result as a future work.

\\
\section{Conclusion}
\label{sec:conclusions}
In this paper, we have presented BigText-QA---a question answering system that uses a large-scale hybrid knowledge graph as its knowledge base. To this end, BigText-QA outperforms DrQA, a state-of-the-art neural-network-based QA system, and achieves competitive results with QUEST, a graph-based unsupervised QA system that inspired the design of BigText-QA. We see these results as a proof-of-concept for our hybrid knowledge representation which captures both textual and structured components in a unified manner in our BigText-KG approach.



\section*{Acknowledgments}
This work was funded by FNR (Grant ID: 15748747).
We thank Rishiraj Saha Roy and his group at the Max Planck Institute for Informatics for their helpful discussions and their support on integrating QUEST with our BigText graph.

\bibliographystyle{splncs04}
\bibliography{anthology,custom}
%
%
%
%




\end{document}